\documentclass[runningheads]{llncs}
\usepackage{graphicx}
\usepackage{amsmath,amssymb} 
\usepackage{epsfig}
\usepackage{color}
\usepackage{booktabs} 
\usepackage{multirow}
\usepackage{subcaption}
\usepackage[font={footnotesize}]{caption}
\usepackage[pagebackref=true,breaklinks=true,letterpaper=true,colorlinks,bookmarks=false]{hyperref}
\usepackage{xspace}
\usepackage{xcolor}

\usepackage[breaklinks=true,bookmarks=false]{hyperref}
\hypersetup{
    colorlinks,
    linkcolor={red!90!black},
    citecolor={green!90!black},
    urlcolor={blue!90!black}
}

\makeatletter
\DeclareRobustCommand\onedot{\futurelet\@let@token\@onedot}
\def\@onedot{\ifx\@let@token.\else.\null\fi\xspace}

\def\eg{\emph{e.g}\onedot}

\def\etc{\emph{etc}\onedot} 
 
\def\etal{\emph{et al}\onedot}
\makeatother

\begin{document}
\title{Instance-level Human Parsing via Part Grouping Network} 

\titlerunning{Instance-level Human Parsing via Part Grouping Network}
%
\author{Ke Gong\inst{1} \and
Xiaodan Liang\inst{1} \thanks{The corresponding author is Xiaodan Liang.}\and
Yicheng Li\inst{1} \and
Yimin Chen\inst{2} \and
Ming Yang\inst{3} \and
Liang Lin\inst{1,2}}
%
\authorrunning{Ke Gong, Xiaodan Liang, Yicheng Li, Yimin Chen, Ming Yang, Liang Lin}
%

\institute{Sun Yat-sen University \and SenseTime Group (Limited) \and CVTE Research \\
\email{gongk3@mail2.sysu.edu.cn, \{xdliang328,liyicheng199506\}@gmail.com, chenyimin@sensetime.com, yangming@cvte.com, linlng@mail.sysu.edu.cn}
}
\maketitle              
\begin{abstract}
Instance-level human parsing towards real-world human analysis scenarios is still under-explored due to the absence of sufficient data resources and technical difficulty in parsing multiple instances in a single pass. Several related works all follow the ``parsing-by-detection" pipeline that heavily relies on separately trained detection models to localize instances and then performs human parsing for each instance sequentially. Nonetheless, two discrepant optimization targets of detection and parsing lead to suboptimal representation learning and error accumulation for final results. In this work, we make the first attempt to explore a detection-free Part Grouping Network (PGN) for efficiently parsing multiple people in an image in a single pass. Our PGN reformulates instance-level human parsing as two twinned sub-tasks that can be jointly learned and mutually refined via a unified network: 1) semantic part segmentation for assigning each pixel as a human part (\eg, face, arms); 2) instance-aware edge detection to group semantic parts into distinct person instances. Thus the shared intermediate representation would be endowed with capabilities in both characterizing fine-grained parts and inferring instance belongings of each part. Finally, a simple instance partition process is employed to get final results during inference. We conducted experiments on PASCAL-Person-Part dataset and our PGN outperforms all state-of-the-art methods. Furthermore, we show its superiority on a newly collected multi-person parsing dataset (CIHP) including 38,280 diverse images, which is the largest dataset so far and can facilitate more advanced human analysis. The CIHP benchmark and our source code are available at \url{http://sysu-hcp.net/lip/}. 

\end{abstract}
\section{Introduction}

Human parsing for recognizing each semantic part (\eg, arms, legs) is one of the most fundamental and critical tasks in analyzing human in the wild and plays an important role in higher level application domains, such as video surveillance~\cite{wang2014deformable}, human behavior analysis~\cite{gan2016concepts,liang2015proposal}. 

Driven by the advance of fully convolutional networks (FCNs)~\cite{long2014fully}, human parsing, or semantic part segmentation has recently witnessed great progress thanks to deeply learned features~\cite{simonyan2014very,he2015deep}, large-scale annotations~\cite{DBLP:journals/corr/LinMBHPRDZ14,Gong_2017_CVPR}, and advanced reasoning over graphical models~\cite{crfasrnn,chen2016deeplab}. However, previous approaches only focus on the single-person parsing task in the simplified and limited conditions, such as fashion pictures~\cite{Yamaguchiparsing13,Dongparsing13,ATR,Co-CNN,chen2014detect} with upright poses and diverse daily images~\cite{Gong_2017_CVPR}, and disregard more real-world cases where multiple person instances appear in one image.  Such ill-posed single-person parsing task severely prohibits the potential applications of human analysis towards more challenging scenarios (\eg, group behavior prediction). 

\begin{figure*}[t]
\centering
  \includegraphics[width=0.9\linewidth]{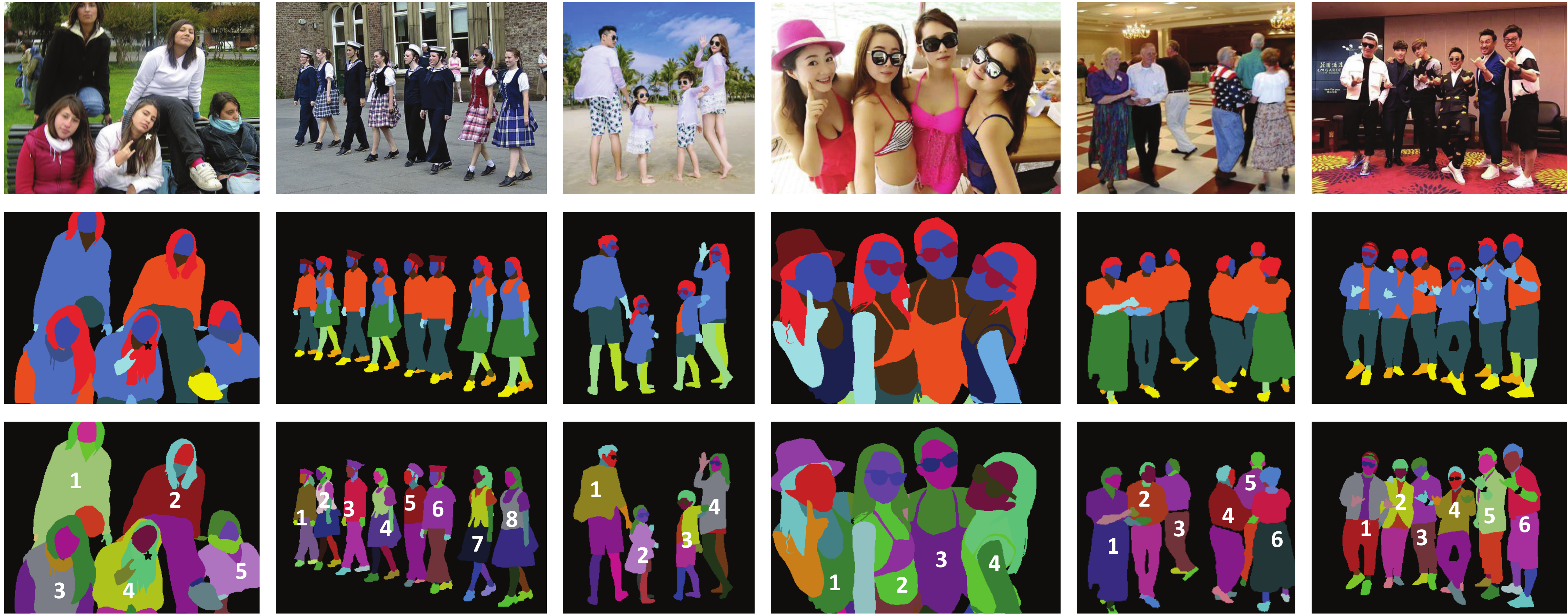}
\vspace{-2mm}
\caption{Examples of our large-scale ``Crowd Instance-level Human Parsing (CIHP)'' dataset, which contains 38,280 multi-person images with elaborate annotations and high appearance variability as well as complexity. The images are presented in the first row. The annotations of semantic part segmentation and instance-level human parsing are shown in the second and third row respectively. Best viewed in color.}
\vspace{-6mm}
\label{fig:cihp}
\end{figure*}

In this work, we aim at resolving the more challenging instance-level human parsing task, which needs to not only segment various body parts or clothes but also associate each part with one instance, as shown in Fig.~\ref{fig:cihp}. Besides the difficulties shared with single-person parsing (\eg, various appearance/viewpoints, self-occlusions), instance-level human parsing is posed as a more challenging task since the number of person instances in an image varies immensely, which cannot be conventionally addressed using traditional single-person parsing pipelines with fixed prediction space that categorizes a fixed number of part labels.

The very recent work~\cite{li2017holistic} explored this task following the ``parsing-by-detection" pipeline~\cite{hariharan2014simultaneous,liang2016reversible,Dai_2016_CVPR,pinheiro2015learning,He_2017_ICCV} that firstly localizes bounding boxes of instances and then performs fine-grained semantic parsing for each box. However, such complex pipelines are trained using several independent targets and stages for the detection and segmentation, which may lead to inconsistent results for coarse localization and pixel-wise part segmentation. For example, segmentation models may predict semantic part regions outside the detected boxes by detection models since their intermediate representations are dragged into different directions.

In this work, we reformulate the instance-level human parsing from a new perspective, that is, tackling two coherent segment grouping goals via a unified network, including the part-level pixel-grouping and instance-level part-grouping. First, part-level pixel-grouping can be addressed by the semantic part segmentation task that assigns each pixel as one part label, which learns the categorization property. Second, given a set of independent semantic parts, instance-level part-grouping can determine the instance belongings of all parts according to the predicted instance-aware edges, where parts that are separated by instance edges will be grouped into distinct person instances. We call this detection-free unified network that jointly optimizes semantic part segmentation and instance-aware edge detection as Part Grouping Network (PGN) illustrated in Fig.~\ref{fig:PGN}. 

\begin{figure}[t]
\begin{center}
   \includegraphics[width=0.9\linewidth]{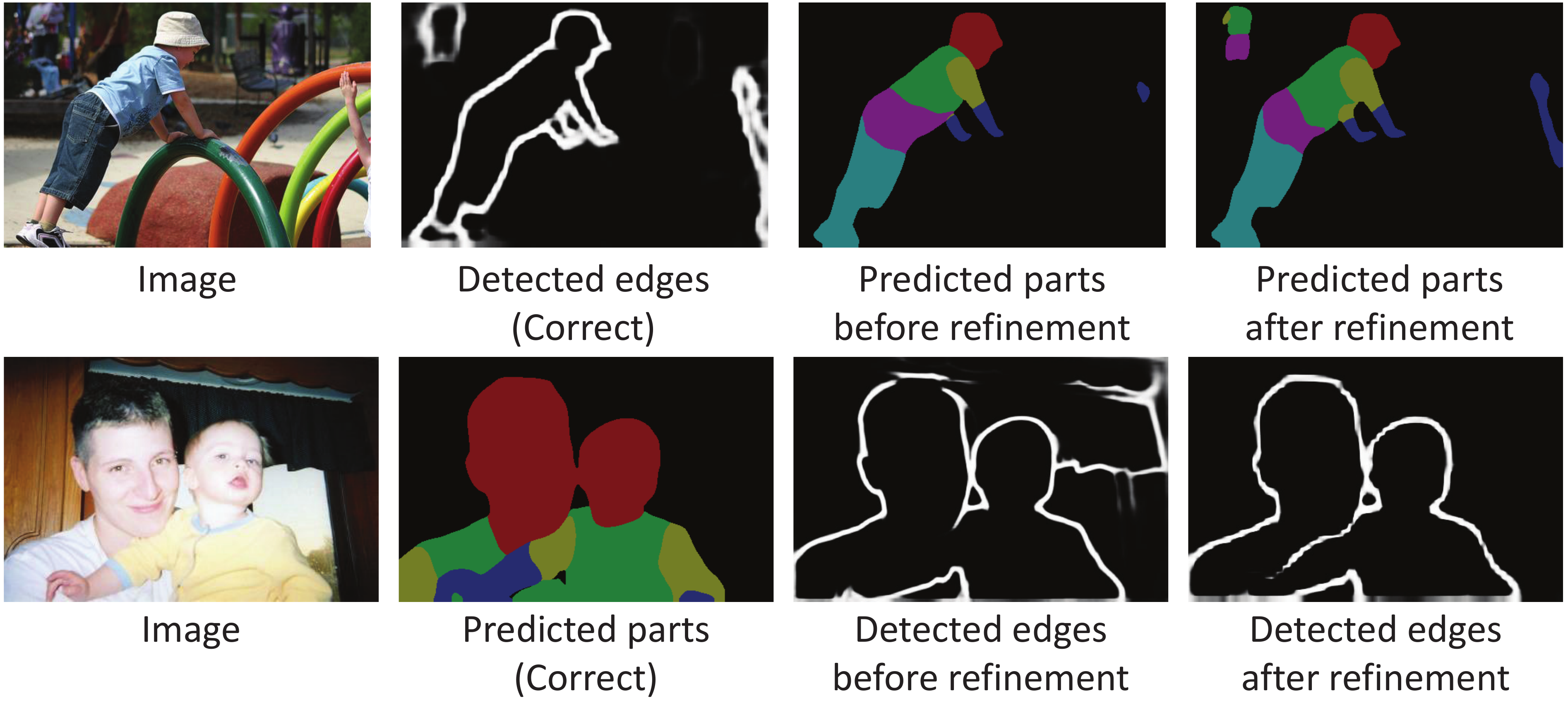}
\end{center}
\vspace{-6mm}
\caption{Two examples show that the errors of parts and edges of challenging cases can be seamlessly remedied by the refinement scheme in our PGN. In the first row, segmentation branch fails to locate the small objects (\eg, the person at the left-top corner and the hand at the right-bottom corner) but edge branch detects them successfully. In the second row, the background edges are mistakenly labeled. However, these incorrect results are rectified by the refinement branch in our PGN.}
\vspace{-6mm}
\label{fig:refined}
\end{figure}

Moreover, unlike other proposal-free methods~\cite{Liu_2017_ICCV,Kirillov_2017_CVPR,liang2015proposal} that break the task of instance object segmentation into several sub-tasks by a few separate networks and resort to complex post-processing, our PGN seamlessly integrates part segmentation and edge detection under a unified network that first learns shared representation and then appends two parallel branches with respect to semantic part segmentation and instance-aware edge detection. As two targets are highly correlated with each other by sharing coherent grouping goals, PGN further incorporates a refinement branch to make two targets mutually benefit from each other by exploiting complementary contextual information. This integrated refinement scheme is especially advantageous for challenging cases by seamlessly remedying the errors from each target. As shown in Fig.~\ref{fig:refined}, a small person may fail to be localized by segmentation branch but successfully detected by edge branch or the mistakenly labeled background edges from instance boundaries could be corrected with our refinement algorithm. Given semantic part segmentation and instance edges, an efficient cutting inference can be used to generate instance-level human parsing results using a breadth-first search over line segments obtained by jointly scanning the segmentation and edges maps.

Furthermore, to our best knowledge, there is no available large-scale dataset for instance-level human parsing research, until our work fills this gap. We introduce a new large-scale dataset, named as Crowd Instance-level Human Parsing (CIHP), including 38,280 multi-person images with pixel-wise annotations of 19 semantic parts in instance-level. The dataset is elaborately annotated focusing on the semantic understanding of multiple people in the wild, as shown in Fig.~\ref{fig:cihp}. With the new dataset, we also propose a public server benchmark for automatically reporting evaluation results for fair comparison on this topic.

Our contributions are summarized in the following aspects. 1) We investigate a more challenging instance-level human parsing, which pushes the research boundary of human parsing to match real-world scenarios much better. 2) A novel Part Grouping Network (PGN) is proposed to solve multi-person human parsing in a unified network at once by reformulating it as two twinned grouping tasks that can be mutually refined: semantic part segmentation and instance-aware edge detection. 3) We build a new large-scale benchmark for instance-level human parsing and present a detailed dataset analysis. 4) PGN surpasses previous methods for both semantic part segmentation and edge detection tasks, and achieves state-of-the-art performance for instance-level human parsing on both the existing PASCAL-Person-Part~\cite{chen2014detect} and our new CIHP dataset.

\section{Related Work}

\textbf{Human Parsing}
Recently, many research efforts have been devoted to human parsing~\cite{Co-CNN,yamaguchi2012parsing,Yamaguchiparsing13,SimoSerraACCV2014,M-CNN,xia2015zoom,chen2015attention,Gong_2017_CVPR} for advancing human-centric analysis research. For example, Liang \etal.~\cite{Co-CNN} proposed a novel Co-CNN architecture that integrates multiple levels of image contexts into a unified network. Gong \etal~\cite{Gong_2017_CVPR} designed a structure-sensitive learning to enforce the produced parsing results semantically consistent with the human joint structures. However, all these prior works only focus on the relatively simple single-person human parsing without considering the common multiple instance cases in the real world. 

As for current data resources, we summarized the publicly available datasets for human parsing in Table~\ref{tab:dataset_num}. Previous datasets only include very few person instances and categories in one image, and require prior works only evaluate pure part segmentation performance while disregarding their instance belongings. On the contrary, containing 38,280 images, the proposed CIHP dataset is the first and also the most comprehensive dataset for instance-level human parsing to date. Although there exist a few datasets in the vision community that were dedicated to other tasks, \eg, clothes recognition, retrieval~\cite{liuLQWTcvpr16DeepFashion,WhereToBuyItICCV15} and fashion modeling~\cite{simo2015neuroaesthetics}, our CIHP that mainly focuses on instance-level human parsing is the largest one and provides more elaborate dense annotations for diverse images. A standard server benchmark for our CIHP can facilitate the human analysis research by enabling fair comparison among current approaches.

\begin{table}[t]
\centering
\scriptsize
\caption{Comparison among the publicly available datasets for human parsing. For each dataset, we report the number of person instances per image, the total number of images, the separate number of images in training, validation, and test sets as well as the number of part labels including the background.}
\tabcolsep 0.02in 
\begin{tabular}{ccccccc}
\toprule[0.7pt]
    Dataset                                  & \# Instances/image  & \# Total & \# Train  & \# Validation & \# Test   & Categories \\ \hline 
    Fashionista~\cite{yamaguchi2012parsing}  &     1               & 685      &  456      &         -      &    229    & 56         \\
    PASCAL-Person-Part~\cite{chen2014detect} &     2.2             & 3,533    &  1,716    &         -      &    1,817  & 7          \\
    ATR~\cite{Co-CNN}                        &     1               & 17,700   &  16,000   &        700     &    1,000  & 18         \\ 
    LIP~\cite{Gong_2017_CVPR}                &     1               & 50,462   &  30,462   &        10,000  &    10,000 & 20         \\ \hline
    CIHP                                     &     3.4             & 38,280   &  28,280   &        5,000   &    5,000  & 20         \\ 
\toprule[0.7pt]
\end{tabular}
\vspace{-8mm}
\label{tab:dataset_num}
\end{table}

\textbf{Instance-level Object Segmentation}    
Our target is also very relevant to instance-level object segmentation task that aims to predict a whole mask for each object in an image. Most of the prior works~\cite{Dai_2016_CVPR,pinheiro2015learning,hariharan2014simultaneous,liang2016reversible,pinheiro2015learning,He_2017_ICCV} addressed this task by sequentially performance optimizing object detection and foreground/background segmentation. Dai \etal~\cite{Dai_2016_CVPR} proposed a multiple-stage cascade to unify bounding box proposal generation, segment proposal generation, and classification. In~\cite{Arnab_2017_CVPR,li2017holistic}, a CRF is used to assign each pixel to an object detection box by exploiting semantic segmentation maps. More recently, Mask R-CNN~\cite{He_2017_ICCV} extended the Faster R-CNN detection framework~\cite{fasterrcnn} by adding a branch for predicting segmentation masks of each region-of-interest. However, these proposal-based methods may fail to model the interactions among different instances, which is critical for performing more fine-grained segmentation for each instance in our instance-level human parsing.

Nonetheless, some approaches~\cite{liang2015proposal,Kirillov_2017_CVPR,Liu_2017_ICCV,Ren_2017_CVPR,Bai_2017_CVPR,romera2016recurrent} are also proposed to bypass the object proposal step for instance-level segmentation. In PFN~\cite{liang2015proposal}, the number of instances and per-pixel bounding boxes are predicted for clustering to produce instance segmentation. In~\cite{Kirillov_2017_CVPR}, semantic segmentation and object boundary prediction were exploited to separate instances by a complicated image partitioning formulation. Similarly, SGN~\cite{Liu_2017_ICCV} proposed to predict object breakpoints for creating line segments, which are then grouped into connected components for generating object regions. Despite their similar intuition with ours in grouping regions to generate an instance, these two pipelines separately learn several sub-networks and thus obtain final results relying on a few independent steps.

Here, we emphasize this work investigates a more challenging fine-grained instance-level human parsing task that integrates the current semantic part segmentation and instance-level object segmentation tasks. From the technical perspective, we present a novel detection-free Part Grouping Network that unifies and mutually refines two twinned grouping tasks in an end-to-end way: semantic part segmentation and instance-aware edge detection. Without the expensive CRF refinement used in~\cite{li2017holistic}, the final results can then be effortlessly obtained by a simple instance partition process.

\section{Crowd Instance-level Human Parsing Dataset}
To benchmark the more challenging multi-person human parsing task, we build a large-scale dataset called Crowd Instance-level Human Parsing (CIHP) Dataset, which has several appealing properties. First, with 38,280 diverse human images, it is the largest multi-person human parsing dataset to date. Second, CIHP is annotated with rich information of person items. The images in this dataset are labeled with pixel-wise annotations on 20 categories and instance-level identification. Third, the images collected from the real-world scenarios contain people appearing with challenging poses and viewpoints, heavy occlusions, various appearances and in a wide range of resolutions. Some examples are shown in Fig.~\ref{fig:cihp}. With the CIHP dataset, we propose a new benchmark for instance-level human parsing together with a standard evaluation server where the test set will be kept secret to avoid overfitting.

\subsection{Image Annotation}
The images in the CIHP are collected from unconstrained resources like Google and Bing. We manually specify several keywords (\eg, family, couple, party, meeting, \etc) to gain a great diversity of multi-person images. The crawled images are elaborately annotated by a professional labeling organization with well quality control. We supervise the whole annotation process and conduct a second-round check for each annotated image. We remove the unusable images that are of low resolution, image quality, or contain one or no person instance.

In total, 38,280 images are kept to construct the CIHP dataset. Following random selection, we arrive at a unique split that consists of 28,280 training and 5,000 validation images with publicly available annotations, as well as 5,000 test images with annotations withheld for benchmarking purposes.

\begin{figure*}[t]
\begin{subfigure}{0.45\textwidth}
  \includegraphics[width=0.75\linewidth]{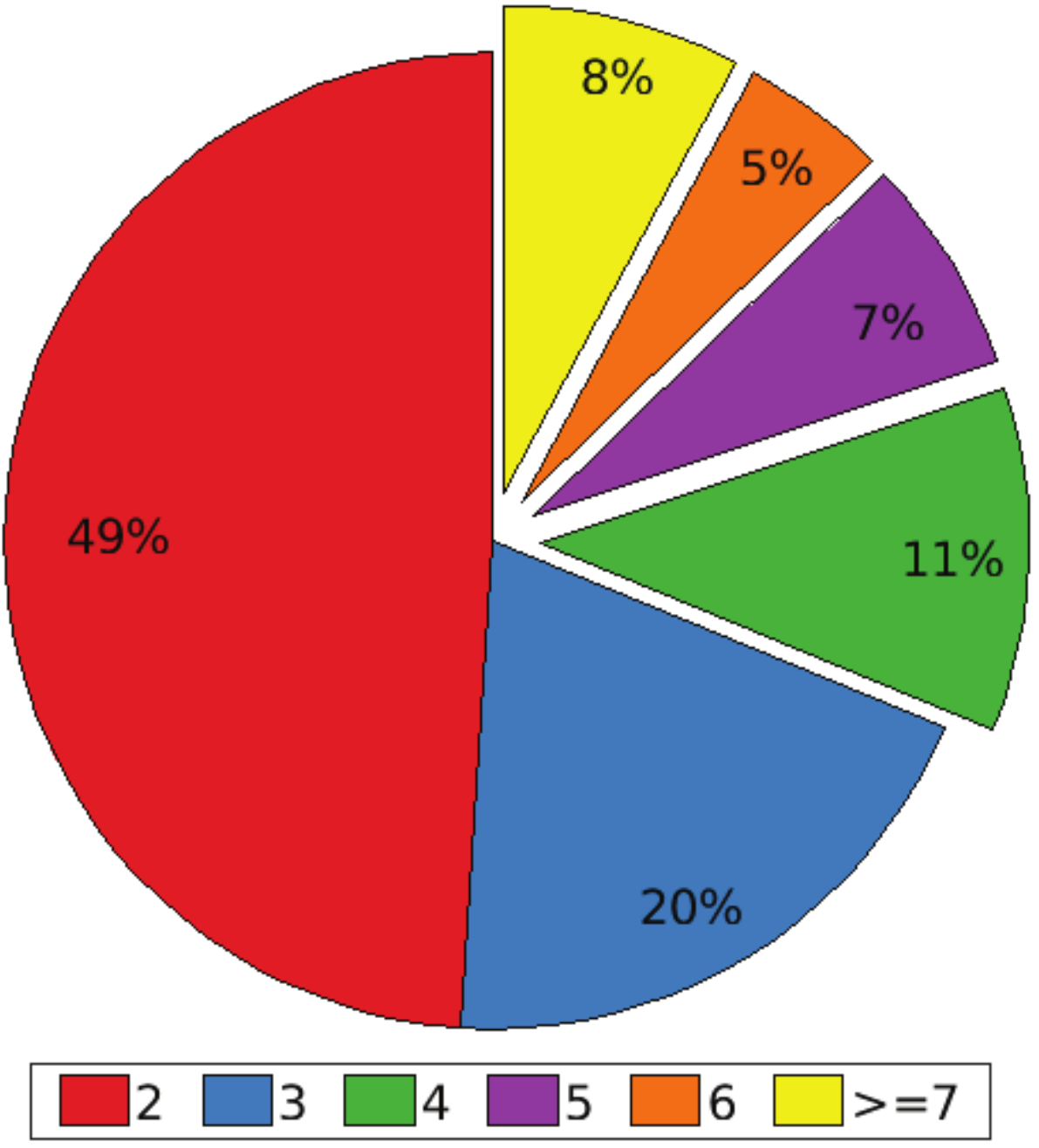}
\end{subfigure}
\begin{subfigure}{0.7\textwidth}
  \includegraphics[width=0.8\linewidth]{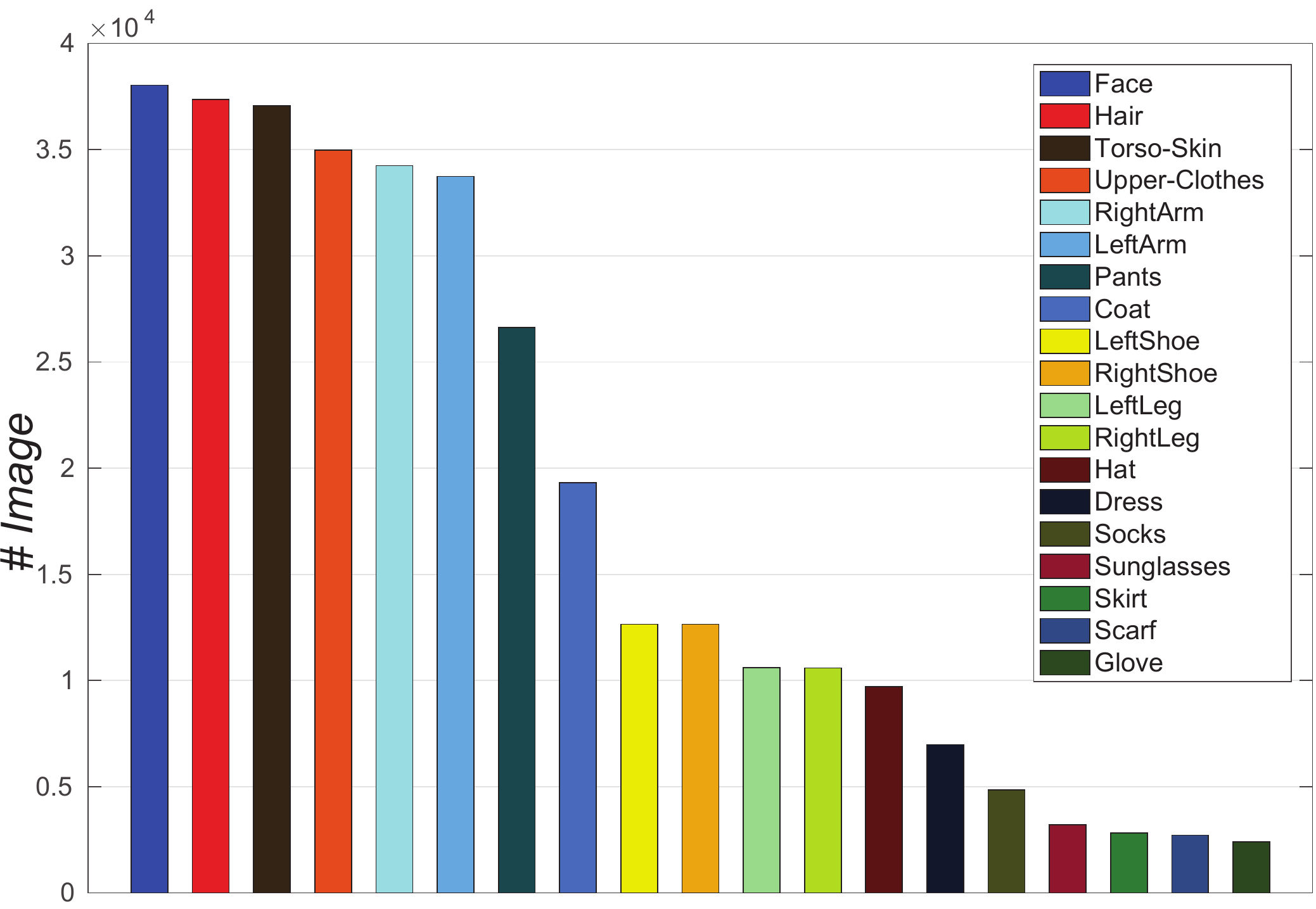}
\end{subfigure}
\vspace{-2mm}
\caption{Left: Statistics on the number of persons in one image. Right: The data distribution on 19 semantic part labels in the CIHP dataset.}
\vspace{-8mm}
\label{fig:num}
\end{figure*}

\subsection{Dataset Statistics}
We now introduce the images and categories in the CIHP dataset with more statistical details. Superior to the previous attempts~\cite{Gong_2017_CVPR,Co-CNN,chen2014detect} with average one or two person instances in an image, all images of the CIHP dataset contain two or more instances with an average of 3.4. The distribution of the number of persons per image is illustrated in Fig.~\ref{fig:num} (Left). Generally, we follow LIP~\cite{Gong_2017_CVPR} to define and annotate the semantic part labels. However, we find that the Jumpsuit label defined in LIP~\cite{Gong_2017_CVPR} is infrequent compared to other labels. To parse the human more completely and precisely, we use a more common body part label (Tosor-skin) instead. The 19 semantic part labels in the CIHP are Hat, Hair, Sunglasses, Upper-clothes, Dress, Coat, Socks, Pants, Gloves, Scarf, Skirt, Torso-skin, Face, Right/Left arm, Right/Left leg, and Right/Left shoe. The numbers of images for each semantic part label are presented in Fig.~\ref{fig:num} (Right).


\section{Part Grouping Network}
In this section, we begin by presenting a general pipeline of our approach (see Fig.~\ref{fig:PGN}) and then describe each component in detail. The proposed Part Grouping Network (PGN) jointly train and refine the semantic part segmentation and instance-aware edge detection in a unified network. Technically, these two sub-tasks are both pixel-wise classification problem, on which Fully Convolutional Networks (FCNs)~\cite{long2014fully} perform well. Our PGN is thus constructed based on FCNs structure, which first learns common representation using shared intermediate layers and then appends two parallel branches with respect to semantic part segmentation and edge detection. To explore and take advantage of the semantic correlation of these two tasks, a refinement branch is further incorporated to make two targets mutually beneficial for each other by exploiting complementary contextual information. Finally, an efficient partition process with a heuristic grouping algorithm can be used to generate instance-level human parsing results using a breadth-first search over line segments obtained by jointly scanning the generated semantic part segmentation maps and instance-aware edge maps.

\subsection{PGN architecture}

\textbf{Backbone sub-network} 
Basically, we use a repurposed ResNet-101 network, Deeplab-v2~\cite{chen2016deeplab} as our human feature encoder, because of its high performance demonstrated in dense prediction tasks. It employs convolution with upsampled filters, or “atrous convolution”, which effectively enlarges the field of view of filters to incorporate larger context without increasing the number of parameters or the amount of computation. The coupled problems of semantic segmentation and edge detection share several key properties that can be efficiently learned by a few shared convolutional layers. Intuitively, they both desire satisfying dense recognition according to low-level contextual cues from nearby pixels and high-level semantic information for better localization. In this way, instead of training two separate networks to handle these two tasks, we perform a single backbone network that allows weight sharing for learning common feature representation. 

\begin{figure*}[t]
\centering
  \includegraphics[width=0.9\linewidth]{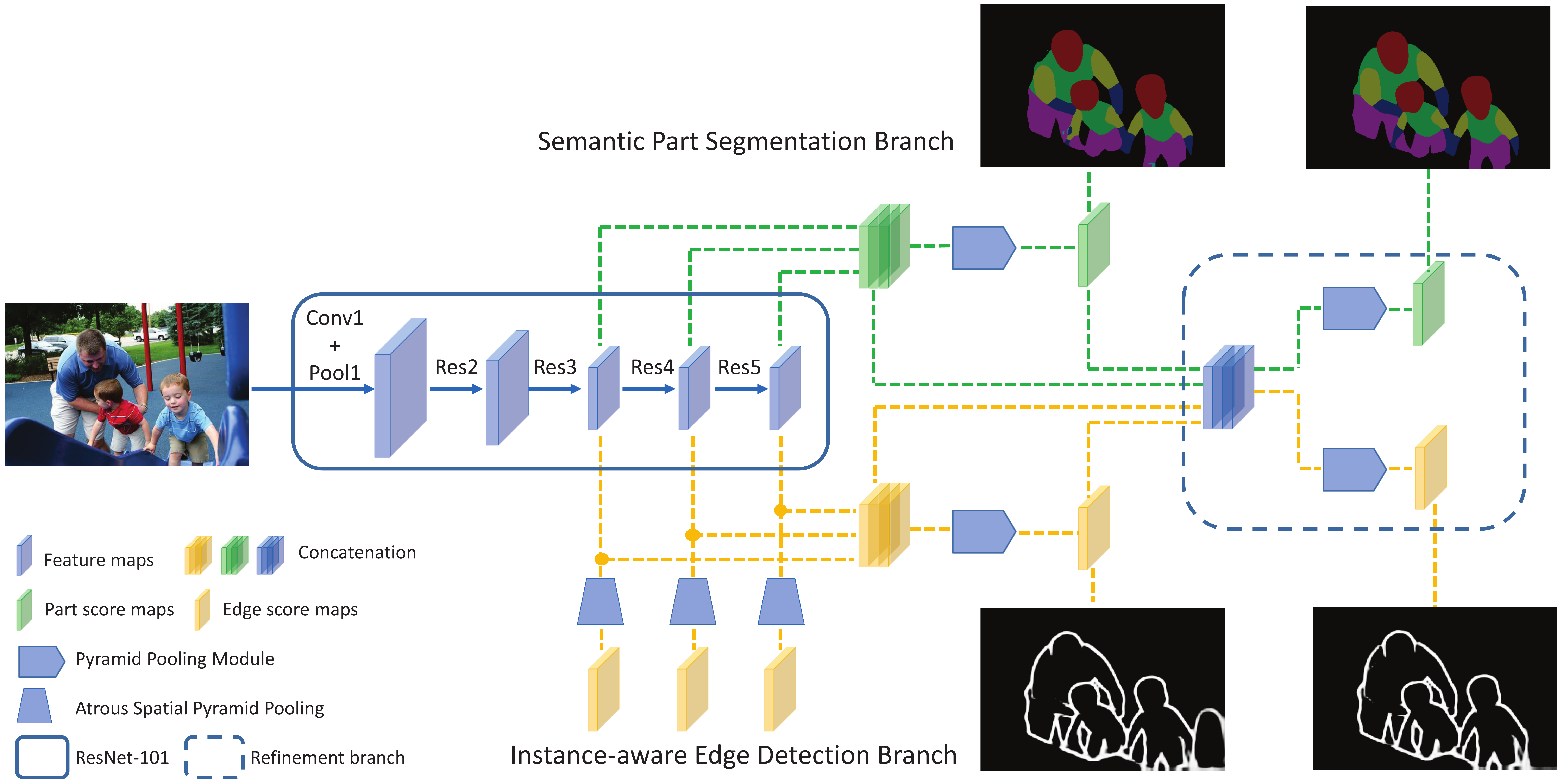}
\vspace{-2mm}
\caption{Illustration of our Part Grouping Network (PGN). Given an input image, we use ResNet-101 to extract the shared feature maps. Then, two branches are appended to capture part context and human boundary context while simultaneously generating part score maps and edge score maps. Finally, a refinement branch is performed to refine both predicted segmentation maps and edge maps by integrating part segmentation and human boundary contexts.}
\vspace{-7mm}
\label{fig:PGN}
\end{figure*}

\begin{figure*}[t]
\begin{center}
  \includegraphics[width=0.9\linewidth]{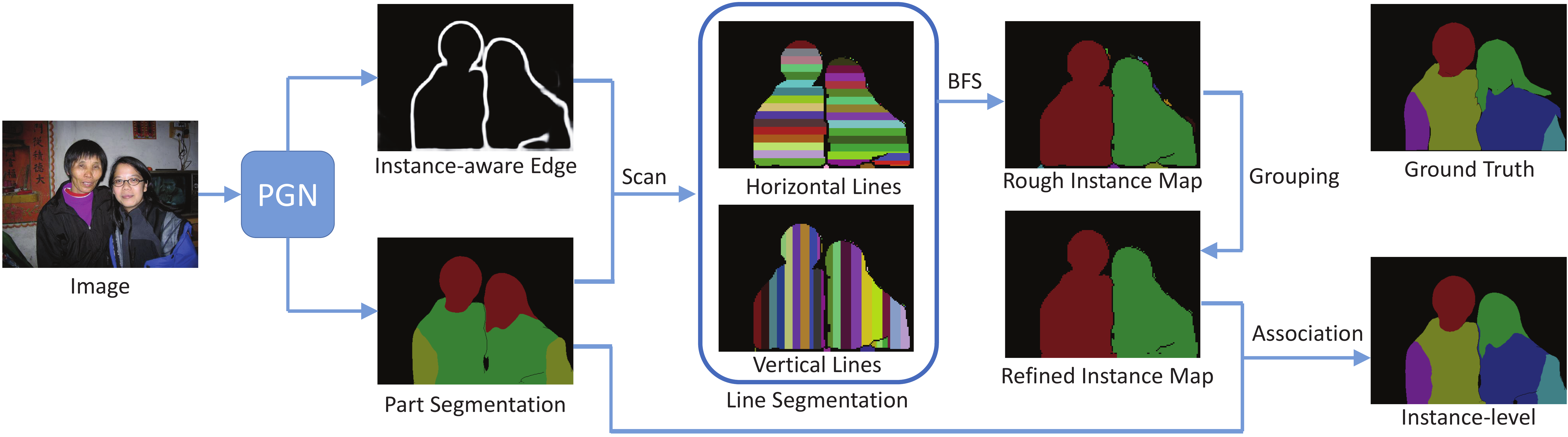}
\end{center}
\vspace{-6mm}
\caption{The whole pipeline of our approach to tackle instance-level human parsing. Generated from the PGN, the part segmentation maps and edge maps are scanned simultaneously to create horizontal and vertical segmented lines. Just like a connected graph problem, the breadth-first search can be applied to group segmented lines into regions. Furthermore, the small regions near the instance boundary are merged into their neighbor regions that cover larger areas and several part labels. Associating the instance maps and part segmentation maps, the pipeline finally outputs a well-predicted instance-level human parsing result without any proposals from object detection.}
\vspace{-6mm}
\label{fig:partition}
\end{figure*}

However, in the original Deeplab-v2 architecture~\cite{chen2016deeplab}, an input image is downsampled by two different ratios (0.75 and 0.5) to produce multi-scale inputs at three different resolutions, which are independently processed by ResNet-101 using shared weights. The output feature maps are then upsampled and combined by taking the element-wise maximum. This multi-scale scheme requires enormous memory and is time-consuming. Alternatively, we use single scale input and employ two more efficient and powerful coarse-to-fine schemes. Firstly, inspired by skip architecture~\cite{long2014fully} that combines semantic information from a deep, coarse layer with appearance information from a shallow, fine layer to produce accurate and detailed segmentation, we concatenate the activations of the final three blocks of ResNet-101 as the final extracted feature maps. Thanks to the atrous convolution, this information combination allows the network to make local predictions instructed by global structure without upscale operation. Secondly, following PSPNet~\cite{Zhao_2017_CVPR} which exploits the capability of global context information by different region-based context aggregation, we use the pyramid pooling module on top of the extracted feature maps before the final classification layers. The extracted feature maps are average-pooled with four different kernel sizes, giving us four feature maps with spatial resolutions $1\times1$, $2\times2$, $3\times3$, and $6\times6$ respectively. Each feature map undergoes convolution and upsampling, before being concatenated together with each other. Benefiting from these two coarse-to-fine schemes, the backbone sub-network is able to capture contextual information with different scales and varying among different sub-regions.

\textbf{Semantic part segmentation branch} 
The common technique~\cite{chen2015attention,chen2016deeplab} for semantic segmentation is to predict the image at several different scales with shared network weights and then combine predictions together with learned attention weights. To reinforce the efficiency and generalization of our unified network, discarding the multi-scale input, we apply another context aggregation pattern with various average-pooling kernel sizes, which is introduced in~\cite{Zhao_2017_CVPR}. We append one side branch to perform pixel-wise recognition for assigning each pixel with one semantic part label. The $1\times1$ convolutional classifiers output $K$ channels, corresponding to the number of target part labels including a background class.

\textbf{Instance-aware edge detection branch}
Following~\cite{xie2015holistically}, we attach side outputs for edge detection to the final three blocks of ResNet-101. Deep supervision is imposed at each side-output layer to learn rich hierarchical representations towards edge predictions. Particularly, we use atrous spatial pyramid pooling (ASPP)~\cite{chen2016deeplab} for the three edge side output layers to robustly detect boundaries at multiple scales. The ASPP we used consists of one $1\times1$ convolution and four $3\times3$ atrous convolutions with dilation rates of 2, 4, 8, and 16. In the final classification layers for edge detection, we use pyramid pooling module to collect more global information for better reasoning. We apply $1\times1$ convolutional layers with one channel for all edge outputs to generate edge score maps.

\textbf{Refinement branch}
We design a simple yet efficient refinement branch for jointly refining segmentation and edge predictions. As shown in Fig.~\ref{fig:PGN}, the refinement branch integrates the segmentation and edge predictions back into the feature space by mapping them to a larger number of channels with an additional $1\times1$ convolution. The remapped feature maps are combined with the extracted feature maps from both the segmentation branch and edge branch, which are finally fed into another two pyramid pooling modules to mutually boost segmentation and edges results.

In summary, the whole learning objective of PGN can be written as:
\begin{equation}
\vspace{-2mm}
L = \alpha \cdot (L_{\text{seg}} + L'_{\text{seg}}) + \beta \cdot (L_{\text{edge}} + L'_{\text{edge}} + \sum \limits_{n=1}^N L_{\text{side}}^n).
\end{equation}
The resolution of the output score maps is $m\times m$, which is the same for both segmentation and edge. So the segmentation branch has a $Km^2$-dimensional output, which encodes $K$ segmentation maps of resolution $m \times m$, one for each of the $K$ classes. During training, we apply a per-pixel softmax and define $L_{\text{seg}}$ as the multinomial cross-entropy loss. $L'_{\text{seg}}$ is the same but for the refined segmentation results. For each $m^2$-dimensional edge output, we use a per-pixel sigmoid binary cross-entropy loss. $L_{\text{edge}}$, $L'_{\text{edge}}$, and $L_{\text{side}}^n$ denote the loss of the first predicted edge, refined edge and the side-output edge respectively. In our network, the number of edge side output, N is 3. $\alpha$ and $\beta$ are the balance weights.

We use the batch normalization parameters provided by ~\cite{chen2016deeplab}, which are fixed during our training process. Our added modules (including ASPP and pyramid pooling module) on top of ResNet eliminate batch normalization because the whole network is trained end-to-end with a small batch size due to the limitation of physical memory on GPU cards. The ReLU activation function is applied following each convolutional layer except the final classification layers.

\subsection{Instance partition process}

Since the couple tasks of semantic part segmentation and instance-aware edge detection are able to incorporate all required information for depicting instance-level human parsing, we thus employ a simple instance partition process to get final results during inference, which groups human parts into instances based on edge guidance. The whole process is illustrated in Fig.~\ref{fig:partition}.

First, inspired by the line decoding process in~\cite{Liu_2017_ICCV}, we simultaneously scan part segmentation maps and edge maps thinned by non-maximal suppression~\cite{xie2015holistically} to create horizontal and vertical line segments. To create horizontal lines, we slide from left to right along each row. The background positions of segmentation maps are directly skipped and a new line starts when we hit a foreground label of segmentation. The lines are terminated when we hit an edge point and a new line should start at the next position. 
We label each new line with an individual number, so the edge points can cut off the lines and produce a boundary between two different instances. We perform similar operations but slide from top to bottom to create vertical lines.

The next step is to aggregate these two kinds of lines to create instances. We can treat the horizontal lines and vertical lines jointly as a connected graph. The points in the same lines can be thought as connected since they have the same labeled number. We traverse the connected graph by the breadth-first search to find connected components. In detail, when visiting a point, we search its connected neighbors horizontally and vertically and then push them into the queue that stores the points belonging to the same regions. As a result, the lines of the same instance are grouped and different instance regions are separated.

This simple process inevitably introduces errors if there are false edge points inside instances, resulting in many small regions at the area around instance boundaries. We further design a grouping algorithm to handle this issue. Rethinking of the separated regions, if a region contains several semantic part labels and covers a large area, it must be a person instance. On the contrary, if a region is small and only contains one part segmentation labels, we can certainly judge it as an erroneously separated region and then merge it to its neighbor instance region. We treat a region as a person instance if it contains at least two part labels and covers an area over 30 pixels, which works best in our experiments. 

Following this instance partition process, person instance maps could be generated directly from semantic part segmentation and instance-aware edge maps. 

\begin{table*}[t]
\centering
\small
\caption{Comparison of semantic part segmentation performance with the state-of-the-art methods on the PASCAL-Person-Part~\cite{chen2014detect}.}
\begin{tabular}{ccccccccc}
\toprule[0.7pt]
   Method                                         &  head  &  torso  &  u-arms &  l-arms &  u-legs &  l-legs &  Bkg   &  Avg    \\ \hline
   HAZN~\cite{xia2015zoom}                        & 80.79  &  59.11  &  43.05  &  42.76  &  38.99  &  34.46  &  93.59 &  56.11  \\  
   Attention~\cite{chen2015attention}             & 81.47  &  59.06  &  44.15  &  42.50  &  38.28  &  35.62  &  93.65 &  56.39  \\ 
   LG-LSTM~\cite{liang2015semantic}               & 82.72  &  60.99  &  45.40  &  47.76  &  42.33  &  37.96  &  88.63 &  57.97  \\     
   LIP~\cite{Gong_2017_CVPR}                      & 83.26  &  62.40  &  47.80  &  45.58  &  42.32  &  39.48  &  94.68 &  59.36  \\ 
   Graph LSTM~\cite{liang2016semantic}            & 82.69  &  62.68  &  46.88  &  47.71  &  45.66  &  40.93  &  94.59 &  60.16  \\
   Structure-evolving LSTM~\cite{Liang_2017_CVPR} & 82.89  &  67.15  &  51.42  &  48.72  &  51.72  & \textbf{45.91} & \textbf{97.18} &  63.57  \\
   DeepLab v2~\cite{chen2016deeplab}              &   -    &     -   &   -     &    -    &   -     &    -    &   -    &  64.94  \\
   Holistic~\cite{li2017holistic}                 &   -    &     -   &   -     &    -    &   -     &    -    &   -    &  66.3   \\\hline 
   PGN (segmentation)                             & 89.98  &  73.70  & 54.75   &  60.26  &  50.58  &  39.16  &  95.09 &  66.22  \\
   PGN (w/o refinement)                           & 90.11  &  72.93  & 54.01   &  59.47  &  54.57  &  42.03  &  95.12 &  66.91  \\\hline 
   \textbf{PGN}                               & \textbf{90.89}  &  \textbf{75.12}  &  \textbf{55.83}  &  \textbf{64.61}  &  
                                                  \textbf{55.42}  &  41.57  &  95.33 &  \textbf{68.40}   \\
\toprule[0.7pt]
\end{tabular}
\vspace{-10mm}
\label{tab: pascal}
\end{table*}

\section{Experiments}

\subsection{Experimental Settings}

\textbf{Training details:}
We use the basic structure and network settings provided by Deeplab-v2~\cite{chen2016deeplab}. The $512 \times 512$ inputs are randomly cropped from the images during training. The size of the output scope maps, $m$ equals to 64 with the downsampling scale of 1/8. The number of category $K$ is 7 for PASCAL-Person-part dataset~\cite{chen2014detect} and 20 for our CIHP dataset.

The initial learning rate is 0.0001, the parsing loss weight $\alpha$ is 1 and the edge loss weight $\beta$ is 0.01. Following~\cite{chen2017rethinking}, we employ a `poly' learning rate policy where the initial learning rate is multiplied by 
$(1-\frac{ \text{iter} }{ \text{max\_iter}})^\text{power}$ with $\text{power} = 0.9$. We train all models with a batch size of 4 images and momentum of 0.9.

We apply data augmentation, including randomly scaling the input images (from 0.5 to 2.0), randomly cropping and randomly left-right flipping during training for all datasets. As reported in~\cite{li2017holistic}, the baseline methods, Holistic~\cite{li2017holistic} and MNC~\cite{Dai_2016_CVPR} are pre-trained on Pascal VOC Dataset~\cite{everingham2010pascal}. For fair comparisons, we train the PGN at the same settings for roughly 80 epochs.

Our method is implemented by extending the TensorFlow framework. All networks are trained on four NVIDIA GeForce GTX 1080 GPUs. The code and models are available at \url{https://github.com/Engineering-Course/CIHP_PGN}.

\textbf{Inference:}
During testing, the resolution of every input is consistent with the original image. We average the predictions produced by the part segmentation branch and the refinement branch as the final results for part segmentation. For edge detection, we only use the results of the refinement branch. To stabilize the predictions, we perform inference by combining results of multi-scale inputs and left-right flipped images. In particular, the scale is 0.5 to 1.75 in increments of 0.25 for segmentation and from 1.0 to 1.75 for edge detection. In partition process, we break the lines when the activation of edge point is larger than 0.2.

\textbf{Evaluation metric:}
The standard intersection over union (IoU) criterion is adopted for evaluation on semantic part segmentation, following~\cite{chen2014detect}. To evaluate instance-aware edge detection performance, we use the same measures for traditional edge detection~\cite{Liu_2017_CVPR}: fixed contour threshold (ODS) and per-image best threshold (OIS). In terms of instance-level human parsing, we define metrics drawing inspirations from the evaluation of instance-level semantic segmentation. Specifically, we adopt mean Average Precision, referred to as $AP^r$~\cite{hariharan2014simultaneous}. We also compare the mean of the $AP^r$ score for overlap thresholds varying from 0.1 to 0.9 in increments of 0.1, noted as $AP^r_\text{vol}$~\cite{li2017holistic}.

\begin{table}[t]
\centering
\small
\caption{Comparison of instance-aware edge detection performance on the PASCAL-Person-Part dataset~\cite{chen2014detect}.}
\tabcolsep 0.25in 
\begin{tabular}{ccc}
\toprule[0.7pt]
   Method                                       & ODS            & OIS   \\ \hline 
   RCF~\cite{Liu_2017_CVPR}                     & 38.2           & 39.8   \\
   CEDN~\cite{yang2016object}                   & 38.9           & 40.1      \\ 
   HED~\cite{xie2015holistically}               & 39.6           & 41.3    \\  \hline
   PGN (edge)                                   & 41.8           & 43.0         \\
   PGN (w/o refinement)                         & 42.1           & 43.5         \\  \hline
   \textbf{PGN}                                 & \textbf{42.5}  & \textbf{43.9}      \\
\toprule[0.7pt]
\end{tabular}
\vspace{-8mm}
\label{tab: edge_result}
\end{table}

\subsection{PASCAL-Person-Part Dataset}

We first evaluate the performance of our PGN on the PASCAL-Person-part dataset~\cite{chen2014detect} with 1,716 images for training and 1,817 for testing. Following~\cite{chen2015attention,xia2015zoom}, the annotations are merged to include six person part classes and one background class which are Head, Torse, Upper/Lower arms and Upper/Lower legs.

\begin{table}[t]
\centering
\small
\caption{Comparison of $AP^r$ at various IoU thresholds for instance-level human parsing on the PASCAL-Person-Part dataset~\cite{chen2014detect}.}
\tabcolsep 0.15in 
\begin{tabular}{ccccc}
\toprule[0.7pt]
\multirow{2}{*}{Method}        & \multicolumn{3}{c}{IoU threshold}   &  \multirow{2}{*}{$AP^r_\text{vol}$}   \\     
                               & 0.5           & 0.6           &  0.7          &                  \\ \hline
MNC~\cite{Dai_2016_CVPR}       & 38.8          & 28.1          & 19.3          & 36.7             \\
Holistic~\cite{li2017holistic} & \textbf{40.6} & \textbf{30.4} & 19.1          & 38.4             \\  \hline
PGN (edge + segmentation)      & 36.2          & 25.9          & 16.3          & 35.6             \\
PGN (w/o refinement)           & 39.1          & 29.3          & 19.5          & 37.8             \\  
PGN (w/o grouping)             & 37.1          & 28.2          & 19.3          & 38.2             \\
PGN (large-area grouping)      & 37.6          & 28.7          & 19.7          & 38.6             \\     \hline
\textbf{PGN}                   & 39.6          & 29.9          & \textbf{20.0} & \textbf{39.2}    \\
\toprule[0.7pt]
\end{tabular}
\vspace{-8mm}
\label{tab: instance_pascal}
\end{table}

\textbf{Comparison on Semantic Part Segmentation} 
We report the semantic part segmentation results compared with the state-of-the-art methods in Table~\ref{tab: pascal}. The proposed PGN substantially outperforms all baselines in terms of most of the categories. Particularly, our best model achieves 2.1\% improvements in average IoU compared with the closest competitor. This superior performance confirms the effectiveness of our unified network on semantic part segmentation, which incorporates the information of object boundaries into the pixel-wise prediction.

\textbf{Comparison on Instance-aware Edge Detection} 
We report the statistic comparison of our PGN and state-of-the-art methods on instance-aware edge detection in Table~\ref{tab: edge_result}. Our PGN gives a huge boost in terms of ODS and OIS. This large improvement demonstrates that edge detection can benefit from semantic part segmentation in our unified network.

\textbf{Comparison on Instance-level Human Parsing} 
Table~\ref{tab: instance_pascal} shows the comparison results of instance-level human parsing with two baseline methods~\cite{Dai_2016_CVPR,li2017holistic}, which rely on object detection framework to generate a large number of proposals for separating instances. Our PGN method achieves state-of-the-art performance, especially in terms of high IoU threshold, thanks to the more smooth boundaries of segmentation refined by edge context. It verifies the rationality of our PGN based on the assumption that semantic part segmentation and edge detection together can directly depict the key characteristics to achieve good capability in instance-level human parsing. The joint feature learning scheme in PGN also makes the part-level grouping by semantic part segmentation and instance-level grouping by instance-aware edge detection mutually benefit from each other by seamlessly incorporating multi-level contextual information.

\begin{table}[t]
\centering
\small
\caption{Performance comparison of edges (Left), part segmentation (Middle) and instance-level human parsing (Right) from different components of PGN on the CIHP.}
\tabcolsep 0.03in 
\begin{tabular}{c|cc|c|cccc}
\toprule[0.7pt]
\multirow{2}{*}{Method}        &  \multirow{2}{*}{ODS} & \multirow{2}{*}{OIS} & \multirow{2}{*}{Mean IoU} 
                               & \multicolumn{3}{c}{IoU threshold} &  \multirow{2}{*}{$AP^r_\text{vol}$}   \\ 
                                &   &  &  & 0.5 & 0.6 & 0.7 &   \\ \hline                     
PGN (edge) + PGN (segmentation) & 44.8       & 44.9      & 50.7   & 28.5  & 22.9  & 16.4  & 27.8  \\
PGN (w/o refinement)            & 45.3       & 45.6      & 54.1   & 33.3  & 26.3  & 18.5  & 31.4  \\   
PGN (w/o grouping)              & -          & -         & -      & 34.7  & 27.8  & 20.1  & 32.9  \\
PGN (large-area grouping)       & -          & -         & -      & 35.1  & 28.2  & 20.4  & 33.4  \\     \hline
\textbf{PGN}   & \textbf{45.5} & \textbf{46.0} & \textbf{55.8} & \textbf{35.8} & \textbf{28.6} & \textbf{20.5} & \textbf{33.6} \\   
\toprule[0.7pt]
\end{tabular}
\vspace{-8mm}
\label{tab: results_cihp}
\end{table}

\subsection{CIHP Dataset}
As there are no available codes of baseline methods~\cite{li2017holistic}, we extensively evaluate each component of our PGN architecture on the CIHP test set, as shown in Table~\ref{tab: results_cihp}. For part segmentation and instance-level human parsing, the performance on CIHP is worse than those on PASCAL-Person-Part~\cite{chen2014detect}, because the CIHP dataset contains more instances with more diverse poses, appearance patterns and occlusions, which is more consistent with real-world scenarios, as shown in Fig.~\ref{fig:visual}. However, the images in CIHP are high-quality with higher resolutions, which makes the results of edge detection become better.

\subsection{Ablation Studies}
We further evaluate the effect of the main components of our PGN.

\textbf{The unified network} We train two independent networks, PGN(segmentation) and PGN(edge), with only a segmentation branch or an edge branch, as reported in Table~\ref{tab: pascal},~\ref{tab: edge_result},~\ref{tab: instance_pascal},~\ref{tab: results_cihp}. From the comparisons, we can learn that our unified network incorporating information from part context and boundaries context can predict a better result than using a single task network. Moreover, the joint training can also improve the final instance-level human parsing results. 

\textbf{The refinement branch} The comparisons between PGN and PGN (w/o refinement) show that our refinement branch helps part segmentation and instance edges benefit each other by exploiting complementary contextual information, which is an implicit joint optimization just like graphical models. With the well-predicted segmentation and edges, our partition algorithm can generate instance-level results more efficiently than other complex decoding processes~\cite{li2017holistic,Liu_2017_ICCV,Arnab_2017_CVPR}.

\textbf{The grouping algorithm} Finally, we prove that the grouping algorithm in the instance partition process is an effective way to refine results of instance-level human parsing, by inspecting the performance of PGN(w/o grouping) in Table~\ref{tab: instance_pascal},~\ref{tab: results_cihp}. Additionally, PGN(large-area grouping) represents that in the grouping algorithm, whether a region is a person instance only depends on if it covers a large area. The results indicate that our proposed framework including the heuristic grouping algorithm can be generalized and works as well in the case of standard instance segmentation where the part labels are not predicted.

\subsection{Qualitative Results}
\begin{figure*}[t]
\centering
  \includegraphics[width=0.95\linewidth]{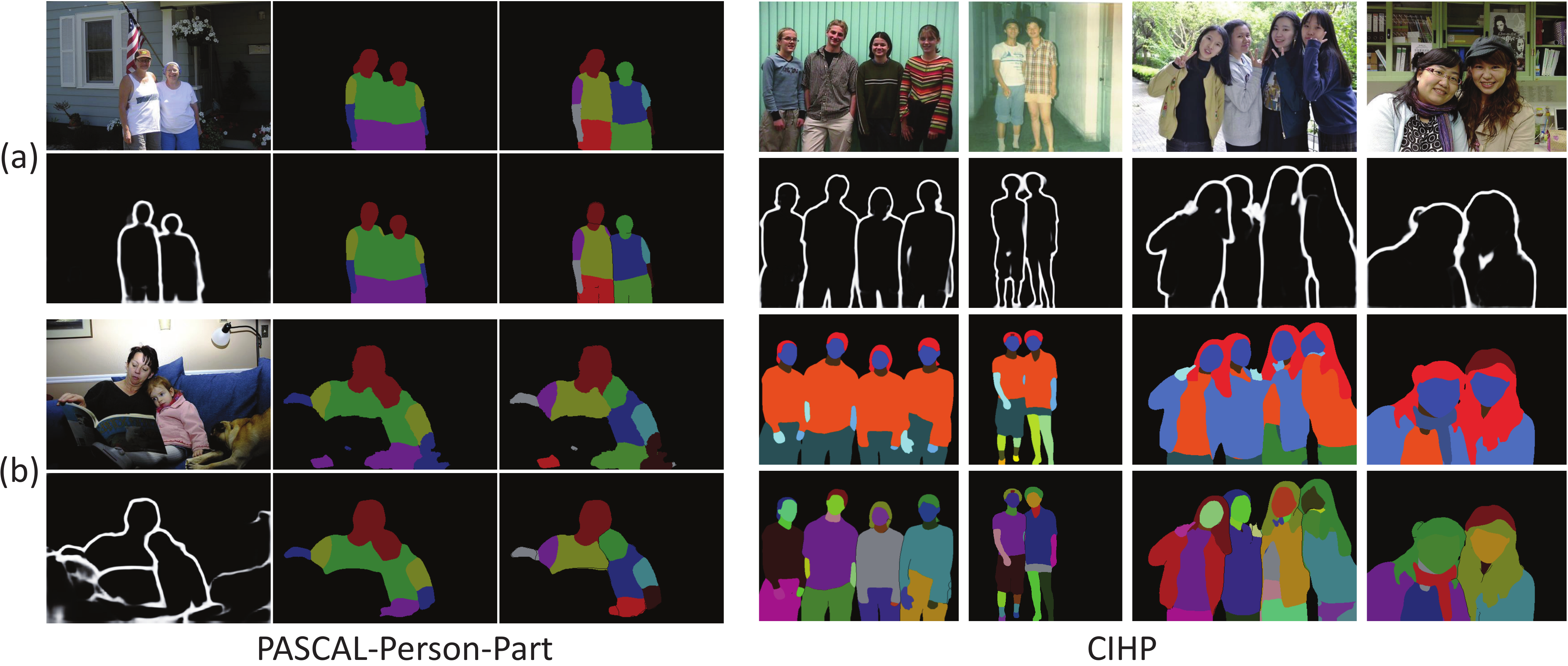}
\vspace{-2mm}
\caption{Left: Visualized results on the PASCAL-Person-Part dataset~\cite{chen2014detect}. In each group, the first line shows the input image, segmentation and instance results of Holistic~\cite{li2017holistic} (provided by the authors), and the results of our PGN are presented in the second line. Right: The images and the predicted resutls of edges, segmentation and instance-level human parsing by our PGN on the CIHP dataset are presented vertically.}
\vspace{-6mm}
\label{fig:visual}
\end{figure*}

The qualitative results on the PASCAL-Person-Part dataset~\cite{chen2014detect} and the CIHP dataset are visualized in Fig.~\ref{fig:visual}. Compared to Holistic~\cite{li2017holistic}, our part segmentation and instance-level human parsing results are more precise because the predicted edges can eliminate the interference from the background, such as the flag in group (a) and the dog in group (b). Overall, our PGN outputs very semantically meaningful predictions, thanks to the mutual refinement between edge detection and semantic part segmentation.

\section{Conclusion}
In this paper, we presented a novel detection-free Part Grouping Network to investigate instance-level human parsing, which is a more pioneering and challenging work in analyzing human in the wild. Our approach jointly optimizes semantic part segmentation and instance-aware edge detection in an end-to-end way and makes these two correlated tasks mutually refine each other. To push the research boundary of human parsing to match real-world scenarios much better, we further introduce a new large-scale benchmark for instance-level human parsing task, including 38,280 images with pixel-wise annotations on 19 semantic part labels. Experimental results on PASCAL-Person-Part~\cite{chen2014detect} and our CIHP dataset demonstrate the superiority of our proposed approach, which surpasses previous methods for both semantic part segmentation and edge detection tasks, and achieves state-of-the-art performance for instance-level human parsing.

\bibliographystyle{splncs04}
\bibliography{egbib}
\end{document}